\documentclass[11pt]{article}
\usepackage{arxiv}
\usepackage[T1]{fontenc}
\usepackage[utf8]{inputenc}
\usepackage{times}
\usepackage{microtype}
\usepackage{placeins}
\usepackage[round,authoryear]{natbib}

\usepackage{amsmath,amssymb}

\usepackage[hidelinks]{hyperref}
\usepackage{url}
\usepackage{xltabular}

\usepackage{xcolor}
\usepackage{colortbl}
\usepackage{booktabs}
\usepackage{array}
\usepackage{threeparttable}
\usepackage{tabularx}
\usepackage{pifont}
\usepackage{graphicx}
\usepackage{subcaption}
\definecolor{UEInk}{HTML}{283846}
\definecolor{UEMuted}{HTML}{788591}
\definecolor{UEBlue}{HTML}{85ABE3}
\definecolor{UEMint}{HTML}{79BEA7}
\definecolor{UERose}{HTML}{DB94B0}
\definecolor{UEBlueCheck}{HTML}{537FB7}
\definecolor{UEMintCheck}{HTML}{378B70}
\definecolor{UERoseCheck}{HTML}{B76186}

\usepackage{multirow}
\usepackage{longtable}

\usepackage{xspace}
\newcommand{\method}{\textbf{DecoEdit}\xspace}
\newcommand{\dataset}{\textbf{DecoDataset}\xspace}
\newcommand{\benchmark}{\textbf{DecoBench}\xspace}

\title{
DecoEdit: Beyond Flat Control for 3D-Aware Unified Image Editing
}

\author{
Xiaotang Gai\textsuperscript{1,2},
Jinbo Xing\textsuperscript{2},
Zijie Meng\textsuperscript{1},
Jiayi Yi\textsuperscript{1,3},
Jimin Xiao\textsuperscript{1,3}\\[0.35em]
Yixuan Qian\textsuperscript{1,3},
Yu Liu\textsuperscript{2},
Chaojie Mao\textsuperscript{2},
Zuozhu Liu\textsuperscript{1,*}\\[0.8em]
\textsuperscript{1}Zhejiang University\\[0.15em]
\textsuperscript{2}Alibaba Group\\[0.15em]
\textsuperscript{3}University of Illinois at Urbana-Champaign\\[0.6em]
Email:
\href{mailto:xiaotang.23@intl.zju.edu.cn}{\texttt{xiaotang.23@intl.zju.edu.cn}}\\[0.3em]
\textsuperscript{*}Corresponding author
}
\date{}

\renewcommand{\shorttitle}{DecoEdit: 3D-Aware Unified Image Editing}
\renewcommand{\headeright}{Preprint}
\makeatletter
\renewcommand{\@maketitle}{
  \vbox{
    \hsize\textwidth
    \linewidth\hsize
    \centering
    \@toptitlebar
    {\LARGE\bfseries\@title\par}
    \@bottomtitlebar
    \vspace{0.7em}
    {\normalfont\large\@author\par}
    \vspace{1.2em}
  }
}
\makeatother

\begin{document}

\maketitle

\begin{abstract}
While recent advances in multimodal image editing have achieved impressive results, current frameworks remain largely confined to ``flat" 2D controls (e.g., text prompts, bounding boxes, or masks). Precise 3D-aware geometric manipulation—such as
coordinating an object's location, physical scale, and 3D orientation while strictly
preserving its fine-grained identity—remains an elusive challenge, typically suffering from severe detail degradation or requiring expensive per-instance fine-tuning.
In this paper, we present \textbf{DecoEdit}, an all-in-one image editing framework that goes beyond flat control by seamlessly integrating 1D textual, 2D spatial, and 3D volumetric inputs under a unified architecture. First, to power this unified model, we establish a data curation pipeline and curate a large-scale dataset, namely \textbf{DecoDataset}, containing 1.6 million samples, encompassing diverse multi-dimensional (1D/2D/3D) tasks and their complex compositions.
Furthermore, we establish a comprehensive image editing benchmark named \textbf{DecoBench} to systematically evaluate multi-task editing performance.
Finally, to address the fundamental trade-off between complex spatial transformations
and identity preservation, we propose a novel decoupled pipeline: \textbf{DecoEdit} factorizes
the editing process by utilizing structural guidance from SAM-3D to anchor the target
geometry, scale, and spatial orientation, while concurrently leveraging a detail-preservation
path to lock in high-fidelity textures.
Extensive experiments
demonstrate that \textbf{DecoEdit} achieves state-of-the-art results. Notably, on the challenging
3D spatial pose and orientation control tasks, \textbf{DecoEdit} significantly outperforms
leading closed-source models such as Seedream 5.0 in both geometric
accuracy and identity consistency.
\end{abstract}

\section{Introduction}

Multimodal image editing has witnessed rapid progress, empowering models to synthesize visually compelling content guided by natural language instructions~\citep{ye2026imgedit,luo2026editscore,ju2026editverse,yin2026reasonedit,han2026unireditbench,mao2026wan,huang2025diffusion,feng2025dit4edit,yu2025anyedit,barrett2002object,hui2024hq}.
However, translating user intent into reliable edits in real-world scenarios remains fundamentally bounded by controllability.
Real-world editing requirements rarely exist in flat isolation: consider an instruction such as ``relocate this wooden chair to the left desk and rotate it
$45^\circ$ toward the camera". This instruction inherently entangles a 1D semantic relocation operation, 2D spatial destination, and a 3D volumetric novel-view rotation and constraints.
  While humans conceive editing effortlessly across these dimensional layers, contemporary image editors~\citep{cao2025hunyuanimage,bagel,labs2025flux,brooks2023instructpix2pix} remain predominantly confined to ``flat" 2D interactions, struggling to faithfully execute 3D-aware manipulations.

Existing literature largely tackles these requirements along fragmented trajectories. Mainstream instruction-guided and subject-driven editors operate on 2D planes, utilizing text prompts, bounding boxes, or binary masks to insert or restyle objects~\citep{brooks2023instructpix2pix,zhao2024ultraedit,labs2025flux,zhang2023magicbrush}.
When requested to perform out-of-plane geometric modifications, these models either fail to perceive spatial depth, distort physical scale, or break visual integrity. Conversely, 3D-aware editing and novel-view synthesis paradigms either rely on expensive per-instance optimization, struggle to seamlessly composite objects into complex natural scenes, or suffer from severe detail degradation~\citep{michel2023object,chen2025blenderfusion,yenphraphai2024image,zhang2025freeinsert,qin2026scenedesigner,zhu2025training}.
These performance bottlenecks stem from two critical factors. On the data side, there is a severe deficiency of large-scale, unified datasets that co-annotate 1D semantic instructions, 2D spatial regions, and 3D geometric poses in natural environments (Tab.~\ref{tab:dataset_comparison}). Existing training corpora either consist purely of flat 2D image pairs or isolated single-object 3D scans. Lacking joint multi-dimensional supervision, existing models are unable to learn generalized cross-dimensional priors, forcing many 3D-aware approaches to regress into expensive per-instance fine-tuning. On the architectural side, a critical root cause lies in the entanglement of appearance and geometry within a single 2D reference: directly rotating or warping an object in pixel space often obliterates fine-grained instance identities—such as intricate textures, embossed logos, and non-rigid structural nuances.

\providecommand{\UEDSTableFont}{\footnotesize}
\providecommand{\UEDSBenchFont}{\footnotesize}
\providecommand{\UEDSOverviewFraction}{0.55}
\providecommand{\UEDSBenchFraction}{0.425}
\ifdefined\UEDSBenchWidth\else\newlength{\UEDSBenchWidth}\fi
\ifdefined\UEDSOverviewWidth\else\newlength{\UEDSOverviewWidth}\fi
\providecommand{\UEDSSetup}{  \definecolor{UEDSBlue}{HTML}{85ABE3}  \definecolor{UEDSMint}{HTML}{79BEA7}  \definecolor{UEDSRose}{HTML}{DB94B0}  \definecolor{UEDSMuted}{HTML}{657180}  \setlength{\abovecaptionskip}{5pt}  \setlength{\belowcaptionskip}{0pt}}
\providecommand{\UEDSYes}{\textcolor{UEDSMint!48!black}{\ensuremath{\checkmark}}}
\providecommand{\UEDSNR}{\textcolor{UEDSMuted}{\textendash}}
\providecommand{\UEDSCite}[1]{\ifdefined\citep\citep{#1}\else\cite{#1}\fi}
\providecommand{\UEDSTableCaption}[2]{  \begingroup
  \expandafter\def\csname @captype\endcsname{table}  \caption{#1}\label{#2}  \endgroup}

\begin{table}[t]
\centering
\begingroup
\UEDSSetup
\UEDSTableFont
\scriptsize
\setlength{\tabcolsep}{7.4pt}
\renewcommand{\arraystretch}{1.05}
\begin{tabular}{@{}lcccc>{\columncolor{UEDSBlue!9}}c>{\columncolor{UEDSMint!9}}c>{\columncolor{UEDSRose!9}}c>{\columncolor{UEDSRose!14}}c}
\toprule
\textbf{Dataset} & \textbf{Size}
  & \shortstack[l]{\textbf{Tasks}}
  & \textbf{MS}
  & \shortstack{\textbf{2D Spatial}}
  & \shortstack{\textbf{2D Object}}
  & \shortstack{\textbf{3D View}}
  & \shortstack{\textbf{3D Object}
    }
  & \shortstack{\textbf{Joint} \textbf{1D/2D/3D}} \\
\midrule
InstructPix2Pix
  & 313K
  & - & \UEDSNR & \UEDSNR & \UEDSNR & \UEDSNR & \UEDSNR & \UEDSNR \\
\addlinespace[1.5pt]
HQ-Edit
  & 200K
  & - & \UEDSNR & \UEDSNR & \UEDSNR & \UEDSNR & \UEDSNR & \UEDSNR \\
\addlinespace[1.5pt]
Pico-Banana-400K
  & 400K
  & 35 & \UEDSNR & \UEDSNR & \UEDSNR & \UEDSNR & \UEDSNR & \UEDSNR \\
\addlinespace[1.5pt]
Emu Edit
  & 10M
  & 16 & \UEDSNR & \UEDSNR & \UEDSNR & \UEDSNR & \UEDSNR & \UEDSNR \\
\addlinespace[1.5pt]
OmniEdit
  & 1.2M
  & 7  & \UEDSYes & \UEDSNR & \UEDSNR & \UEDSNR & \UEDSNR & \UEDSNR \\
\addlinespace[1.5pt]
MagicBrush
  & 10K
  & - & \UEDSNR & \UEDSYes & \UEDSNR & \UEDSNR & \UEDSNR & \UEDSNR \\
\addlinespace[1.5pt]
UltraEdit
  & 4.1M
  & 9 & \UEDSYes & \UEDSYes & \UEDSNR & \UEDSNR & \UEDSNR & \UEDSNR \\
\addlinespace[1.5pt]
ImgEdit
  & 1.2M
  & 13  & \UEDSYes & \UEDSNR & \UEDSYes & \UEDSNR & \UEDSNR & \UEDSNR \\
\addlinespace[1.5pt]
AnyEdit
  & 2.5M
  & 25 & \UEDSYes & \UEDSYes & \UEDSYes & \UEDSYes & \UEDSNR & \UEDSNR \\
\midrule
\textbf{\dataset (Ours)}
  & \textbf{1.63M}
  & \textbf{42}
  & \UEDSYes & \UEDSYes & \UEDSYes & \UEDSYes & \UEDSYes & \UEDSYes \\
\bottomrule
\end{tabular}
\caption{\textbf{Comparison of image editing datasets}.
Our dataset combines {42 control subtasks}, {650k 3D training samples},
and {joint 1D/2D/3D control} within a unified set.
\textit{MS}: multiple data sources. \textit{2D Spatial}: mask, box or layout. \textit{2D Object}: object image input for reference. \textit{3D View}: explicit rotation/viewpoint. \textit{3D Object}: inserting objects from images into 3D tasks.
}
\label{tab:dataset_comparison}

\endgroup
\end{table}

To overcome these limitations, our work systematically explores unified image editing across three core pillars: \textbf{Data}, \textbf{Evaluation}, and \textbf{Methodology}.
First, we establish a unified data curation pipeline that seamlessly synthesizes 1D textual instructions, 2D signals, and 3D volumetric inputs derived from 3D models. This pipeline scales to over 1.6 million instances and forms \dataset.
It systematically pairs standard textual edit families with fine-grained spatial and geometric conditions, covering both individual controls and their combinatorial intersections under rigorous quality verification.
Second, we introduce \benchmark, a comprehensive evaluation suite designed to assess multi-dimensional image editing. Moving beyond flat 2D metrics, \benchmark comprehensively gauges semantic adherence, spatial placement, and 3D angle precision.
Third, we present \method, an all-in-one editing framework built upon a Mixture-of-Transformers (MoT)~\citep{bagel,liang2024mixture} paradigm that natively ingests heterogeneous  conditional tokens. To resolve the trade-off between structural reorientation and identity preservation, \method adopts an appearance-geometry decoupled pipeline: it leverages normalized structural renderings from SAM-3D~\cite{chen2026sam3d} to explicitly anchor target geometry, scale, and 3D orientation,
while simultaneously routing the uncorrupted source object image to lock in high-frequency instance textures.
Experiments confirm that \method achieves state-of-the-art performance. Crucially, on the challenging 3D orientation control task, \method markedly outperforms top proprietary models such as  Seedream 5.0 in both rotation accuracy and identity consistency.

\section{Related Work}

\textbf{Image Editing.}
Instruction-guided image editing has evolved from synthetic pair generation~\citep{brooks2023instructpix2pix} and human-annotated triplets~\citep{zhang2023magicbrush} to expressive MLLM-directed editing~\citep{fu2024guiding} and large-scale synthetic pipelines~\citep{zhao2024ultraedit}. Concurrently, subject-driven and spatially-grounded editing paradigms leverage reference images or spatial priors (e.g., bounding boxes, masks, and sketches) to insert or replace visual concepts while enforcing localization~\citep{li2023gligen,zhang2023adding}. More recently, unified editing frameworks and generalist models have emerged to handle diverse tasks within integrated architectures~\citep{yu2025anyedit,liu2025step1x,wu2026omnigen2,zhao2026qwen}. While these models broaden semantic and multimodal capabilities, they remain fundamentally anchored in flat 2D operations.

\textbf{3D-Aware Image Manipulation.}
To transcend planar constraints, a growing line of research explores 3D-aware image editing and novel-view manipulation. Early methods lift 2D contents into 3D proxy geometries for manual editing followed by generative refinement~\citep{yenphraphai2024image}, or inject projected geometric keypoints and camera priors~\citep{chen2024geodiffusion,shi2023zero123}. Other works formulate 3D editing via procedural synthetic training~\citep{michel2023object}, specialized graphic engines like Blender~\citep{chen2025blenderfusion}, multi-object 9-DoF coordinate maps~\citep{qin2026scenedesigner}, or multi-stage inpainting and composition cascades~\citep{zhu2025training,zhang2025freeinsert}. Despite impressive spatial manipulations, existing 3D editing solutions often rely on expensive per-instance optimization, multi-step heuristic pipelines, or fail to composite objects seamlessly into complex backgrounds. More critically, they suffer from appearance-geometry entanglement, which degrades high-frequency instance details (e.g., textures, logos) during rotation.

\textbf{Image Editing Datasets and Benchmarks.}
The development of controllable editing is profoundly steered by training corpora and evaluation benchmarks. Early datasets focused on paired instruction-following pairs~\citep{brooks2023instructpix2pix,zhang2023magicbrush}, which were later expanded to million-scale synthesized sets covering diverse textual tasks and spatial regions~\citep{zhao2024ultraedit,sheynin2024emu}. On the evaluation front, benchmarks such as GEdit-Bench~\citep{liu2025step1x} and AnyEdit~\citep{yu2025anyedit} have systematized multi-task visual evaluation. Nevertheless, an evident gap exists in multi-dimensional supervision: current datasets either curate flat 2D modifications or isolate 3D object assets devoid of contextual environments, lacking joint annotations spanning 1D instructions, 2D boxes, and 3D poses. This data scarcity prevents models from learning cross-dimensional priors.

\section{Dataset, Benchmark and Method}
We first describe the large-scale data curation and rigorous filtering pipeline of \dataset (Sec.~\ref{sec:dataset}), followed by the standardized evaluation protocol of \benchmark (Sec.~\ref{sec:benchmark}). Finally, we elaborate on the architecture and training pipeline of our unified editor, \method (Sec.~\ref{sec:method}).

\subsection{DecoDataset: Scalable Multi-Dimensional Data Engine}
\label{sec:dataset}
\begin{figure}[t]
      \centering
      \includegraphics[width=0.95\textwidth]{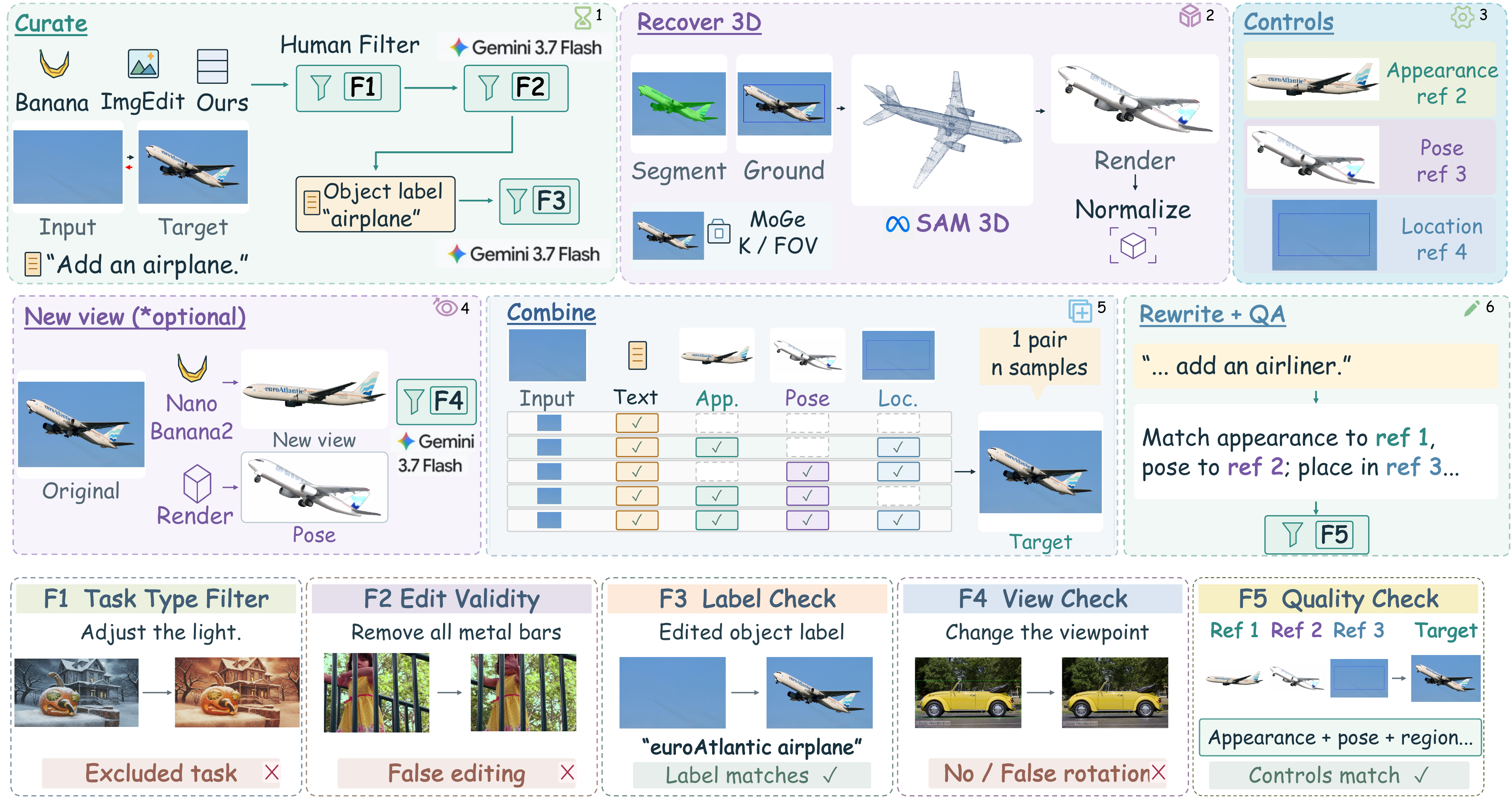}

      \caption{\textbf{Data construction pipeline.}  It involves five filters to select high-quality samples.}
      \label{fig:dataset_pipeline}

\end{figure}

Training a unified multi-dimensional editor requires balanced supervision across text, 2D coordinates, and 3D geometric views. To achieve this at scale, we establish an automated data curation and cascading verification pipeline (Fig.~\ref{fig:dataset_pipeline}). Sourced from open-domain corpora (\textit{ImgEdit}~\citep{ye2026imgedit}, \textit{Banana-400k}~\citep{qian2026pico}) and our in-house collection, \dataset scales ordinary single-turn 2D edits into a multi-control training corpus exceeding 1.6 million instances.

\subsubsection{Data Construction Pipeline}
\label{sec:data_construction}

\textbf{Task Inversion and Semantic Grounding.}
We first identify clean single-turn edits and symmetrically invert them to double data diversity (e.g., inverting \emph{addition} into \emph{removal}, and augmenting bi-directional \emph{replacements}). For each edit pair $(I_{\text{in}}, T_{\text{raw}}, I_{\text{out}})$, Gemini-3.7-Flash automatically detects the targeted entity label. Using this label, we segment the object and ground its 2D bounding box $B^*$, decoupling spatial localization from pixel synthesis.

\textbf{Metric 3D Geometry and Pose Alignment.}
To extract rigid 3D target orientations without manual annotations, we design an automated \emph{reconstruct-render-align} loop:
 \emph{(1) 3D Reconstruction:} Using the extracted foreground mask, SAM-3D~\citep{chen2026sam3d} predicts initial geometry, 3D translation, and rotation, while MoGe~\citep{wang2025moge} estimates monocular metric depth and camera intrinsics $K$.
        \emph{(2) Differentiable Pose Optimization:} To ensure exact alignment with target pixels, we enable layout post-processing: translation and scale are optimized for up to 5 steps, followed by 25 steps of joint pose-scale-translation optimization against target RGB and mask cues. If the resulting mask IoU drops, the optimization safely falls back to the initial pose.
         \emph{(3) Normalized Pose Scaffold ($I_g^*$):} After transforming into a canonical coordinate space, the reconstructed object is rendered via original camera intrinsics. The object is then tightly cropped and scaled to occupy $\approx 85\%$ of a standardized $1024 \times 1024$ white canvas, producing a normalized, geometry-oriented pose reference with coarse appearance $I_g^*$.

\textbf{View Augmentation for Decoupled Training.}
To explicitly empower the model's appearance--geometry decoupling, we synthesize novel-view variations using Gemini-3.1-Image-Preview. Specifically, we generate: (i) \emph{in-the-wild reference edits} with adapted lighting, pose, and background, and (ii) \emph{appearance-isolated novel views} where foreground instances rotate across arbitrary angles while strictly preserving high-frequency textures and background context.

\textbf{Combinatorial Multi-Control Synthesis and Instruction Rewriting.}
A single visual edit pair naturally scales into an entire family of training instances (``$1\text{ pair} \to n\text{ samples}$''). By independently toggling conditions, we assemble combinations: $\{B^*\}$, $\{I_g^*\}$, $\{B^* + I_g^*\}$, with or without appearance references $I_a$. Finally, Gemini-3.7-Flash rewrites the raw edit texts into role-specific instructions (e.g., specifying that the target must strictly adopt the view of reference pose $I_g^*$ while locking the texture of reference $I_a$ into box $B^*$).

\subsubsection{Five-Stage Cascaded Data Filtering}
\label{sec:data_filtering}

To ensure high visual fidelity and precise dimensional alignment, the raw candidates undergo a five-stage automated filtering cascade.
\textbf{F1: Task Type Filter.} We explicitly discard ill-posed manipulation types, such as extreme artistic style transfer (e.g., anime, Van Gogh), blind outpainting, and zoom-in modifications that violate geometric consistency.
 \textbf{F2: Edit Validity.} Gemini-3.7-Flash filters out editing failures, multi-object entanglements, and unintentional background modifications to enforce clean, single-instance supervision.
 \textbf{F3: Label Check.} We verify that the extracted object label strictly matches both the visual pixels and textual instructions, eliminating tracking drift.
  \textbf{F4: View Check.} For augmented novel-view samples, Gemini-3.7-Flash discards cases where rotations failed, background drifted, or physical unnaturalness occurred.
   \textbf{F5: Quality Check.} After full combinatorial assembly and instruction rewriting, all samples undergo final verification evaluated by Gemini-3.7-Flash across 9 rigorous criteria (covering image validities, spatial box compliance, orientation correctness, reference detail fidelity, and unintended human modifications). Only samples achieving a perfect score (5/5) across all nine dimensions are retained, guaranteeing a production-grade training corpus.

\subsection{DecoBench: Multi-Dimensional Evaluation Suite}
\label{sec:benchmark}

To rigorously assess multi-dimensional editing, we construct \benchmark, focusing on real-world user simulation.

\textbf{Cross-Task Sampling.}
For each 3D-related manipulation task, we construct several representative cases. To enable fair and fine-grained cross-task comparisons, we prioritize instances sharing the identical object ID across distinct task branches, extracting all multi-task occurrences of that identity. When shared IDs are exhausted, the remaining quota is filled via uniform random sampling.

\textbf{Strict Zero-Leakage.}
To ensure strict evaluation integrity, we enforce complete data decontamination: once an instance ID is chosen for \benchmark, all corresponding instances in the training set are thoroughly purged—regardless of whether that identity appears in the source canvas, target output, or conditional controls ($B^*, I_a, I_g^*$). This guarantees zero test-set memorization or data leakage.

\textbf{Expert Curation.}
The initial candidates are \textbf{manually constructed} by human experts to ensure unambiguous instructions and balanced visual diversity.
To simulate realistic open-domain user interactions, we augment the benchmark along two lines: (1) We directly manipulate object 3D meshes to render custom, unconstrained viewing angles, closely reflecting arbitrary interactive camera controls from real users, resulting in 256 testing cases. (2) We intentionally include 5 physically violating edge cases (e.g., an object defying gravity or unusual structural orientations) to probe the model's controllability boundaries and constraint adherence under adversarial prompts.
This results in a clean, decontaminated benchmark containing 261 testing cases.

\subsection{DecoEdit: Architecture and Decoupled Methodology}
\label{sec:method}

\begin{figure}[t]
      \centering
      \includegraphics[width=0.86\textwidth]{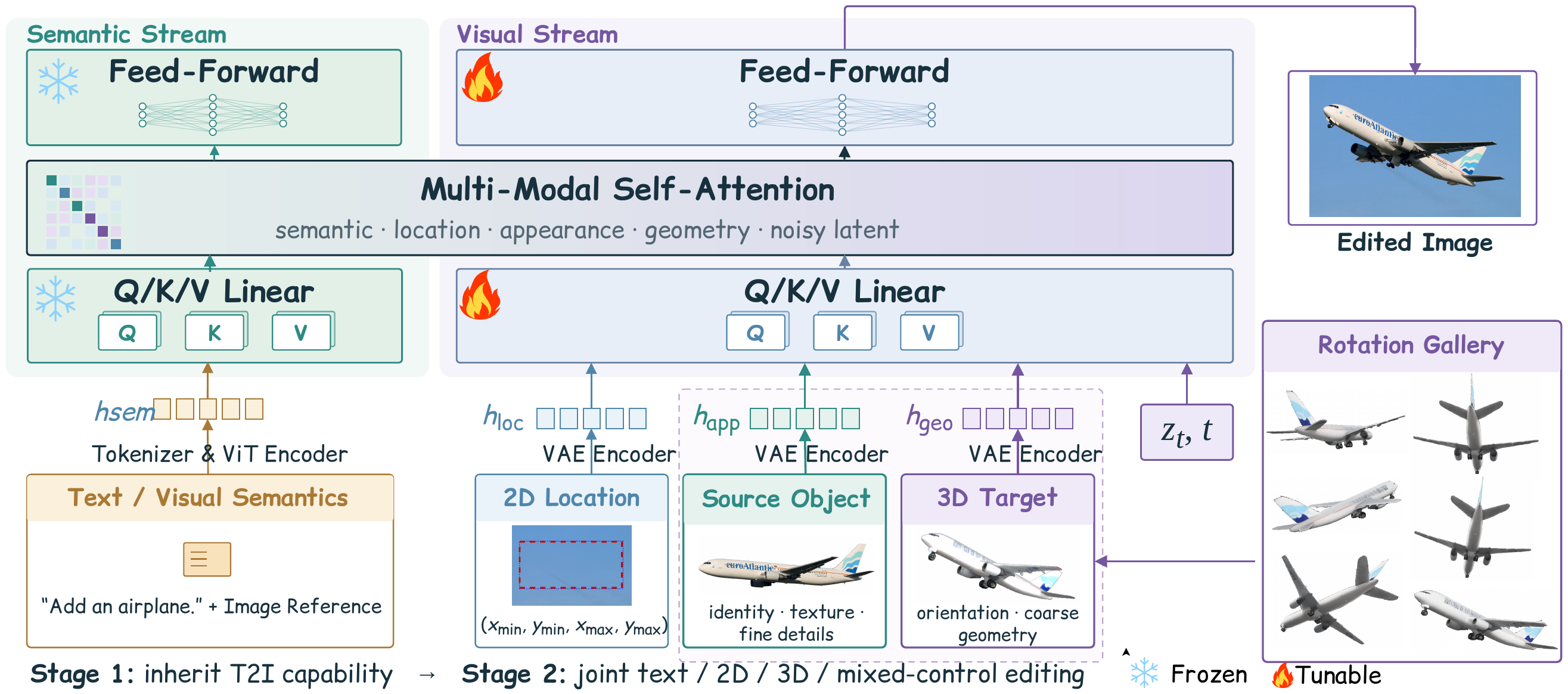}

      \caption{\textbf{The MoT architecture of \method}. It comprises  a semantic stream and a visual stream.}
      \label{fig:model}

\end{figure}

\subsubsection{Overall Framework}

\method unifies heterogeneous conditions into a dual-stream backbone consisting of a {Semantic Stream} and a {Visual Stream}, interconnected via a shared {Multi-Modal Self-Attention} module as illustrated in Fig.~\ref{fig:model}.

\textbf{Input Tokenization.}
The model ingests four role-identifiable conditional inputs:
 \textbf{Semantic Tokens ($h_{\text{sem}}$):} Text instructions and semantic image references $T$ are mapped into semantic embeddings via a Tokenizer and ViT encoder: $h_{\text{sem}} = \mathrm{Encoder}_{\text{sem}}(T)$, respectively.
 \textbf{2D Spatial Tokens ($h_{\text{loc}}$):} Target coordinates $B^* = (x_{\min}, y_{\min}, x_{\max}, y_{\max})$ are converted into visual mask layouts and encoded by the VAE encoder $\mathcal{E}_{\text{vae}}$ into spatial tokens $h_{\text{loc}}$.
 \textbf{Source Appearance Tokens ($h_{\text{app}}$):} The cropped source instance $I_a$ is encoded as $h_{\text{app}} = \mathcal{E}_{\text{vae}}(I_a)$, serving to lock in identity, logos, and high-frequency textures.
 \textbf{3D Target Pose Tokens ($h_{\text{geo}}$):} The SAM-3D geometric rendering $I_g^*$ is encoded into latent geometry tokens $h_{\text{geo}} = \mathcal{E}_{\text{vae}}(I_g^*)$.
 \textbf{Noisy Latent ($z_t$):} The diffusion target latent at timestep $t$ is denoted as $z_t$.

\textbf{MoT Structure.}
The semantic tokens $h_{\text{sem}}$ are routed into dedicated Semantic Q/K/V projections and feed-forward networks (FFN), preserving high-level semantic intent. In parallel, the spatial and conditional visual tokens $H_{\text{vis}} = [h_{\text{loc}} \,\|\, h_{\text{app}} \,\|\, h_{\text{geo}} \,\|\, z_t]$ are projected via a shared Visual Q/K/V linear layer. To enable dynamic cross-dimensional reasoning, the token streams are concatenated along the sequence dimension and interact through Multi-Modal Self-Attention module for information interaction.

\subsubsection{Appearance-Geometry Decoupled Strategy}

\textbf{Limitations of Coupled 3D Conditioning.}
An intuitive approach for 3D-aware image editing is to condition the diffusion model directly on a 3D structural rendering to simultaneously dictate the target orientation and instance details, akin to the paradigm adopted by ControlNet~\citep{zhang2023adding} and its 3D variants.
However, our empirical investigations reveal that existing open-domain 3D reconstruction proxies—including state-of-the-art formulations such as SAM-3D~\citep{chen2026sam3d}—are fundamentally bounded in representational resolution: while they provide coarse volumetric geometry and rigid camera poses, they inevitably discard high-frequency instance textures, fine edge boundaries, and micro-structures (e.g., logos, complex patterns, and materials). Consequently, models following this coupled strategy can satisfy the requested rotation angle, but invariably suffer from severe texture blurring and catastrophic detail degradation.

To resolve the classical trade-off where rotating an object destroys its fine-grained textures, \method establishes an appearance-geometry decoupled editing pipeline: (1) Orientation Guidance via 3D Proxy: rather than forcing the network to infer unseen poses from 2D pixels, we query a pre-rendered \emph{Rotation Gallery} generated via SAM-3D to obtain $I_g^*$. This rendering acts as a pure orientation image, supplying rigid 3D viewpoint and elevation guidance. (2) Detail Preservation via Original Instance Reference: concurrently, the uncorrupted source reference $I_a$ is routed strictly to anchor high-frequency details. Because $I_a$ is relieved from the burden of dictating orientation, the model transfers intricate textures, non-rigid patterns, and text markings from $I_a$ directly onto the geometric framework provided by $I_g^*$.

To achieve balanced multi-task mastery without catastrophic interference, \method is trained in two distinct stages. First, we train our model on broad text-to-image pairs to align the cross-stream projection spaces and inherit open-domain visual priors. In the second stage, we train \method on our curated \dataset for unified editing.

\section{Experiments}

\subsection{Experimental Setup}
\label{Experimental_Setup}

\textbf{Implementation Details.}
Our method instantiates the dual-stream design  as a
MoT backbone with two parameter-disjoint streams of 7B parameters
each a semantic stream and a visual DiT stream  for a total of
14B parameters.
We initialize our model from a pretrained 7B Qwen multimodal backbone~\citep{bai2025qwen3} and fine-tune its visual generation branch, while keeping the language branch, vision encoder, and VAE frozen.
Training uses AdamW with a learning rate of $2.5\times10^{-5}$,
  $\beta_1=0.9$, $\beta_2=0.95$, and weight decay of $0.02$.

\textbf{Evaluation Metrics.}
Instruction-driven editing is intrinsically hard to measure with conventional low-level metrics such as PSNR as documented explicitly in ImgEdit~\citep{ye2026imgedit} and motivates the
judge-based protocols of recent editing benchmarks~\citep{liu2025step1x,yu2025anyedit,
luo2026editscore}.
We therefore evaluate with an MLLM judge (Gemini-3.7-Flash) under a rubric of
seven criteria, and report the seven per-aspect scores numerically.
This principle conditions the seven scored dimensions, each graded on a
$0$--$10$ scale: \textbf{IA} (Instruction Adherence), \textbf{SCA} (Spatial Control Accuracy),
\textbf{PVA} (Pose and View Alignment), \textbf{IAF} (Identity and Appearance Fidelity),
\textbf{NCP} (Non-target Content Preservation), \textbf{IEQ} (Image-Editing Quality), and
\textbf{DCP} (Detail Completion Plausibility).
In parallel, we conduct human evaluation: four annotators,
blind to model identity, rank the outputs of five sampled models per case with ties allowed.

\textbf{Competitors.}
We compare against several open-source methods covering general
generation/editing and unified multimodal families: HiDream-O1-Image(-Dev)~\citep{cai2026hidream}, BAGEL~\citep{bagel},
OmniGen~\citep{xiao2025omnigen}, OmniGen2~\citep{wu2026omnigen2}, FireRed-Image-Edit-1.0/1.1~\citep{team2026firered}, DreamOmni2~\citep{xia2025dreamomni2}, HunyuanImage 3.0~\citep{cao2025hunyuanimage},
Qwen-Image-Edit-2511~\citep{zhao2026qwen}, Qwen-Image-2.1~\citep{zhao2026qwen}, FLUX.2 [klein] 4B/9B~\citep{labs2025flux}, FLUX.2 [dev] and
SenseNova-U1.5-8B-MoT~\citep{diao2026sensenova}. The {closed-source} group covers commercial APIs:
Gemini~2.5~Flash~Image, Gemini~3.1~Flash~Image, Wan~2.7~Image~\citep{mao2026wan}, Seedream~5.0, FLUX.2 [max], and
GPT-Image-1.5.

\subsection{Main Results}
\label{sec:main_results}

\textbf{Quantitative Comparison.}
Tab.~\ref{tab:3d_bench_macro_comparison} summarizes the comparison. \method attains a macro
score of {8.97}, the impressive result among all comparable models, surpassing the strongest
open-source baseline SenseNova-U1.5-8B-MoT ($8.01$) by $+0.96$ and outperforming every leading
closed-source editor: FLUX.2 [max] ($8.84$),
GPT-Image-1.5 ($8.71$), Seedream~5.0 ($8.58$), Gemini~3.1~Flash~Image ($8.58$) and Wan~2.7~Image
($8.53$), while using far fewer parameters than the large open-source competitor
(HunyuanImage~3.0, $80$B total / $13$B active).

\begin{table}[!htbp]
\centering
\begingroup
\scriptsize
\setlength{\tabcolsep}{8.2pt}
\renewcommand{\arraystretch}{1.1}
\definecolor{UEOpenBand}{HTML}{EAF1FA}
\definecolor{UEClosedBand}{HTML}{F1EDF7}
\definecolor{UEOursBand}{HTML}{EAF5EF}
\definecolor{UEGainColor}{HTML}{008A4B}
\definecolor{UEMacroCol}{HTML}{FFDE7A}
\definecolor{UEIACol}{HTML}{EAF2FC}
\definecolor{UESCACol}{HTML}{E3F1F7}
\definecolor{UEPVACol}{HTML}{EEE9FA}
\definecolor{UEIAFCol}{HTML}{E6F3EE}
\definecolor{UENCPCol}{HTML}{EAF5E4}
\definecolor{UEIEQCol}{HTML}{F8EAF1}
\definecolor{UEDCPCol}{HTML}{F5EAF8}
\definecolor{UESafetyCol}{HTML}{EFF1F4}
\newcommand{\UEGainScore}[2]{\shortstack[r]{#1\\[-1.8pt]{\textcolor{UEGainColor}{\fontsize{6.2}{6.2}\selectfont\bfseries #2}}}}
\newcommand{\UEModel}[1]{\mbox{#1}}
\newcommand{\UEThird}[1]{\textit{#1}}
\newcommand{\UEDeemph}[1]{\textcolor{black!52}{#1}}
\begin{threeparttable}
\begin{tabularx}{\linewidth}{@{}>{\raggedright\arraybackslash}X
>{\columncolor{UEIACol}}r
>{\columncolor{UESCACol}}r
>{\columncolor{UEPVACol}}r
>{\columncolor{UEIAFCol}}r
>{\columncolor{UENCPCol}}r
>{\columncolor{UEIEQCol}}r
>{\columncolor{UEDCPCol}}r
>{\columncolor{UEMacroCol}}r
r@{}}
\toprule
\textbf{Model} & \textbf{IA} & \textbf{SCA} & \textbf{PVA} & \textbf{IAF}
& \textbf{NCP} & \textbf{IEQ} & \textbf{DCP} &  \textbf{Macro} & \shortstack{\textbf{DiT Param.}
} \\
\midrule
\rowcolor{UEOpenBand}
\multicolumn{10}{c}{\rule{0pt}{2.1ex}\textbf{Open-source models}} \\
\addlinespace[1.5pt]
\UEModel{HiDream-O1-Image-Dev} & 1.17 & 1.57 & 2.21 & 2.14 & 1.17 & 2.04 & 0.27  & 2.34 & 8B \\
\UEModel{HiDream-O1-Image} & 2.08 & 3.15 & 3.51 & 2.88 & 3.11 & 2.85 & 1.08 & 3.52 & 8B \\
\addlinespace[1pt]
\UEModel{BAGEL} & 2.54 & 3.35 & 4.24 & 3.20 & 3.10 & 2.61 & 1.69  & 3.55 & 7B \\
\addlinespace[1pt]
\UEModel{OmniGen} & 0.53 & 0.78 & 0.85 & 0.71 & 0.97 & 1.48 & 0.29  & 1.26 & 3.8B \\
\UEModel{OmniGen2} & 2.75 & 3.46 & 4.70 & 4.53 & 5.06 & 5.01 & 2.47  & 4.82 & 4B \\
\addlinespace[1pt]
\UEModel{FireRed-Image-Edit-1.0} & 3.28 & 4.20 & 5.17 & 5.32 & 4.78 & 5.55 & 4.13  & 4.68 & 20B \\
\UEModel{FireRed-Image-Edit-1.1} & 4.41 & 5.51 & 5.91 & 6.02 & 7.31 & 6.86 & 4.92  & 5.72 & 20B \\
\addlinespace[1pt]
\UEModel{DreamOmni2} & 4.72 & 6.07 & 5.87 & 5.91 & 7.63 & 7.26 & 5.37  & 6.25 & 12B \\
\addlinespace[1pt]
\UEModel{HunyuanImage 3.0 (API)} & 5.68 & 6.31 & 7.25 & 6.95 & 7.19 & 6.66 & 6.10  & 6.92 & 80A13B \\
\addlinespace[1pt]
\UEModel{Qwen-Image-Edit-2511} & 3.81 & 5.29 & 5.41 & 4.68 & 8.11 & 7.52 & 3.35  & 5.99 & 20B \\
\UEModel{Qwen-Image-2.1} & 6.15 & 6.79 & 7.30 & 7.07 & 8.05 & 6.98 & 5.45  & 6.94 & 7B \\
\addlinespace[1pt]
\UEModel{FLUX.2 [klein] 4B} & 5.05 & 5.94 & 6.29 & 6.02 & 9.21 & 8.46 & 4.98  & 6.81 & 4B \\
\UEModel{FLUX.2 [dev]} & 5.26 & 6.25 & 7.02 & 6.58 & 9.12 & 7.54 & 5.69  & 7.02 & 32B \\
\UEModel{FLUX.2 [klein] 9B} & 6.74 & 7.36 & 7.56 & 7.66 & 9.27 & 8.35 & 6.47  & 7.55 & 9B \\
\addlinespace[1pt]
\UEModel{SenseNova-U1.5-8B-MoT} & 7.75 & 8.16 & 7.92 & 8.30 & 8.36 & 7.65 & 7.75  & 8.01 & 8B \\
\midrule
\rowcolor{UEClosedBand}
\multicolumn{10}{c}{\rule{0pt}{2.1ex}\textbf{Closed-source models}} \\
\addlinespace[1.5pt]
\UEModel{Gemini 2.5 Flash Image} & 6.19 & 7.02 & 7.10 & 7.28 & \underline{9.47} & 8.36 & 6.71  & 7.29 & -- \\
\UEModel{Gemini 3.1 Flash Image} & 8.60 & 8.78 & 8.68 & 8.73 & \UEThird{9.44} & 8.22 & 8.58  & 8.58 & -- \\
\addlinespace[1pt]
\UEModel{Wan 2.7 Image} & 8.45 & 8.79 & 8.59 & 8.68 & 9.42 & 8.36 & 8.50  & 8.53 & -- \\
\addlinespace[1pt]
\UEModel{Seedream 5.0} & \UEThird{8.69} & \UEThird{8.86} & 8.75 & \underline{8.99} & \UEThird{9.44} & 8.67 & 8.77  & 8.58 & -- \\
\addlinespace[1pt]
\UEModel{FLUX.2 [max]} & \underline{8.87} & \underline{8.87} & \underline{9.02} & \UEThird{8.96} & \textbf{9.52} & \textbf{9.15} & \UEThird{8.78}  & \underline{8.84} & -- \\
\addlinespace[1pt]
\UEModel{GPT-Image-1.5} & 8.52 & 8.46 & \UEThird{8.87} & \textbf{9.14} & 8.75 & \underline{9.06} & \textbf{9.18}  & \UEThird{8.71} & -- \\
\midrule
\rowcolor{UEOursBand}
\multicolumn{10}{c}{\rule{0pt}{2.1ex}\textbf{Ours}} \\
\addlinespace[1.5pt]
\UEModel{Baseline} & 5.74 & 6.59 & 6.42 & 6.07 & 9.05 & 7.76 & 5.41  & 6.98 & 7B \\
\UEModel{\method} & \UEGainScore{\textbf{8.98}}{\textbf{+3.24}} & \UEGainScore{\textbf{9.07}}{\textbf{+2.48}} & \UEGainScore{\textbf{9.05}}{\textbf{+2.63}} & \UEGainScore{8.70}{\textbf{+2.63}} & \UEGainScore{9.39}{\textbf{+0.34}} & \UEGainScore{\UEThird{8.92}}{\textbf{+1.16}} & \UEGainScore{\underline{8.90}}{\textbf{+3.50}}  & \UEGainScore{\textbf{8.97}}{\textbf{+1.99}} & 7B \\
\bottomrule
\end{tabularx}

\caption{\textbf{Comparison on \benchmark}.
Refer to Sec.~\ref{Experimental_Setup} for evaluation metric abbreviations. Baseline denotes the initialization checkpoint before fine-tuning. The w/o Decoupling and w/o Check variants are separately trained ablations.
}
\label{tab:3d_bench_macro_comparison}

\end{threeparttable}
\endgroup
\end{table}

\textbf{Human Agreement with The Judge.}
\label{sec:human_agreement}
We verify that the judge is a faithful
proxy for human preference. Four raters independently rank five models (ours, Seedream~5.0,
HunyuanImage~3.0, Gemini~2.5~Flash~Image, Qwen-Image-Edit-2511) on $50$ randomly sampled
\benchmark cases, allowing ties; ranks are normalized to a $0$--$100$ score
(Tab.~\ref{tab:human_scores_vlm_agreement}). Human consensus orders the models as $70.25 > 66.31 > 44.31 > 42.31 >
26.81$, i.e., ours first and Seedream~5.0 close behind.
Agreement between the two is strong and statistically significant: per-rater Spearman
$\rho$ with Gemini-3.7-Flash ranges from $0.451$ to $0.732$ (all adjusted $p<0.001$), and the
per-case mean-rank consensus reaches $\rho=0.733$ with Kendall $\tau_b=0.648$, using $100{,}000$
within-case two-sided permutations and Holm correction across raters.

\begin{table}[!htbp]
\centering
\begingroup
\definecolor{UERaterOne}{HTML}{BCCBDD}
\definecolor{UERaterTwo}{HTML}{79A4B1}
\definecolor{UERaterThree}{HTML}{AA9CC4}
\definecolor{UERaterFour}{HTML}{D7AD7C}
\scriptsize
\setlength{\tabcolsep}{4.7pt}
\renewcommand{\arraystretch}{1.0}
\begin{tabular}{@{}lcccccccc@{}}
\toprule
Rater & \textbf{Ours} & \shortstack{Seedream 5.0} & \shortstack{HY-Image 3.0} & \shortstack{Gemini 2.5 Flash} & \shortstack{Qwen-Image-Edit} & \shortstack{Spearman $\rho\uparrow$} & \shortstack{$p$-value} & \shortstack{Kendall $\tau_b\uparrow$} \\
\midrule
Rater 1 & \textbf{72.50} & \underline{62.75} & 44.50 & \textit{44.75} & 25.50 & 0.644 & $<0.001$ & 0.547 \\
Rater 2 & \underline{61.25} & \textbf{71.25} & \textit{48.25} & 41.25 & 28.00 & 0.451 & $<0.001$ & 0.383 \\
Rater 3 & \textbf{74.25} & \underline{66.00} & 40.00 & \textit{42.25} & 27.50 & 0.676 & $<0.001$ & 0.592 \\
Rater 4 & \textbf{73.00} & \underline{65.25} & \textit{44.50} & 41.00 & 26.25 & 0.732 & $<0.001$ & 0.655 \\
\textbf{Average} & \textbf{70.25} & \underline{66.31} & \textit{44.31} & 42.31 & 26.81 & 0.733 & $<0.001$ & 0.648 \\
\bottomrule
\end{tabular}

\caption{Human evaluation and agreement with the VLM judge on 3D editing tasks.
Model columns report normalized human ranking scores (0--100; higher is better).
Correlation columns report mean within-case correlations with Gemini-3.7-Flash
}
\label{tab:human_scores_vlm_agreement}

\endgroup
\end{table}

\begin{figure}[!htbp]
      \centering
      \includegraphics[width=0.9\textwidth]{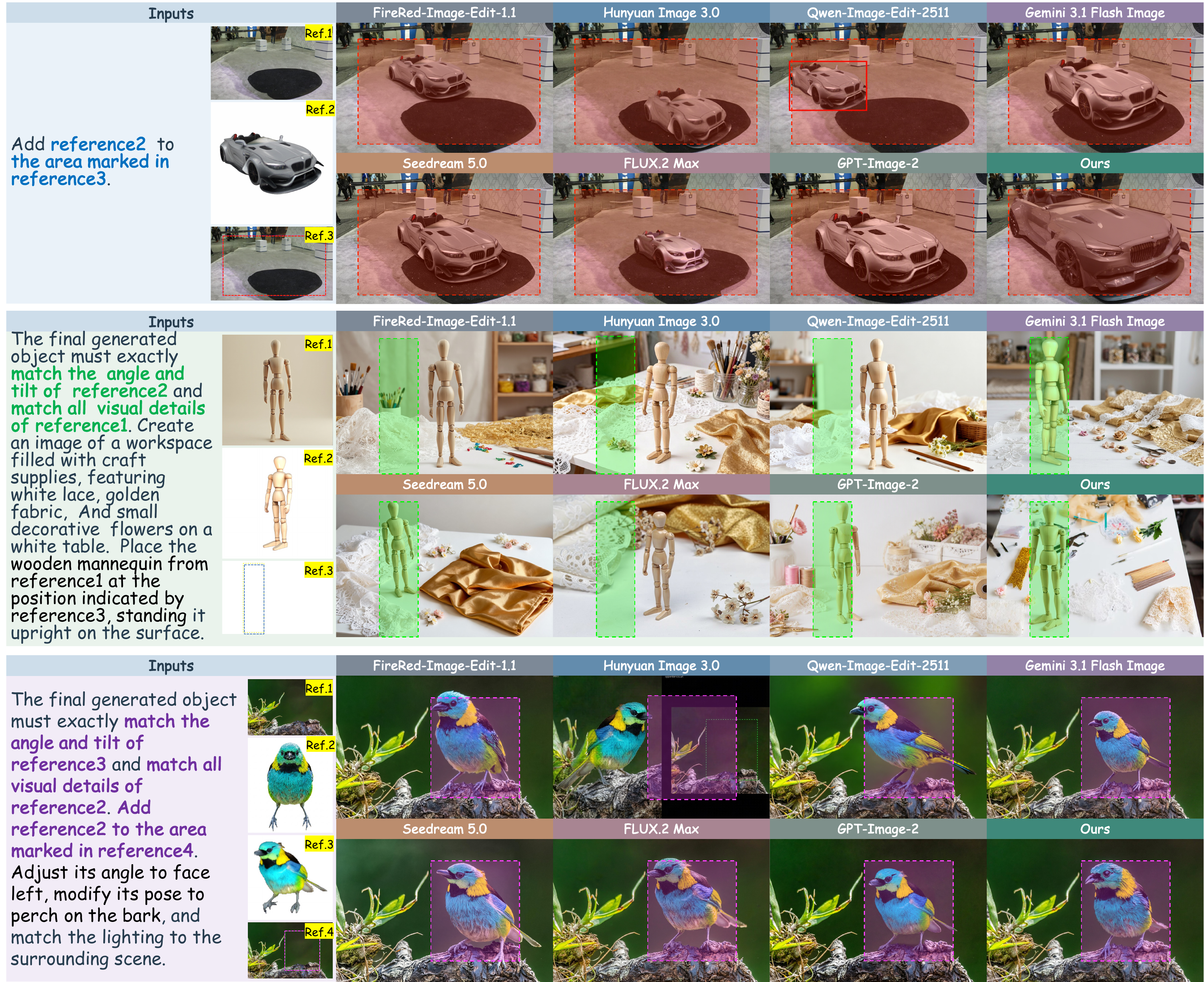}

      \caption{\textbf{Qualitative comparison with existing state-of-the-art method} on 3D-aware tasks.}
      \label{fig:main_exp}

\end{figure}

\begin{figure}[!htbp]
      \centering
      \includegraphics[width=0.9\textwidth]{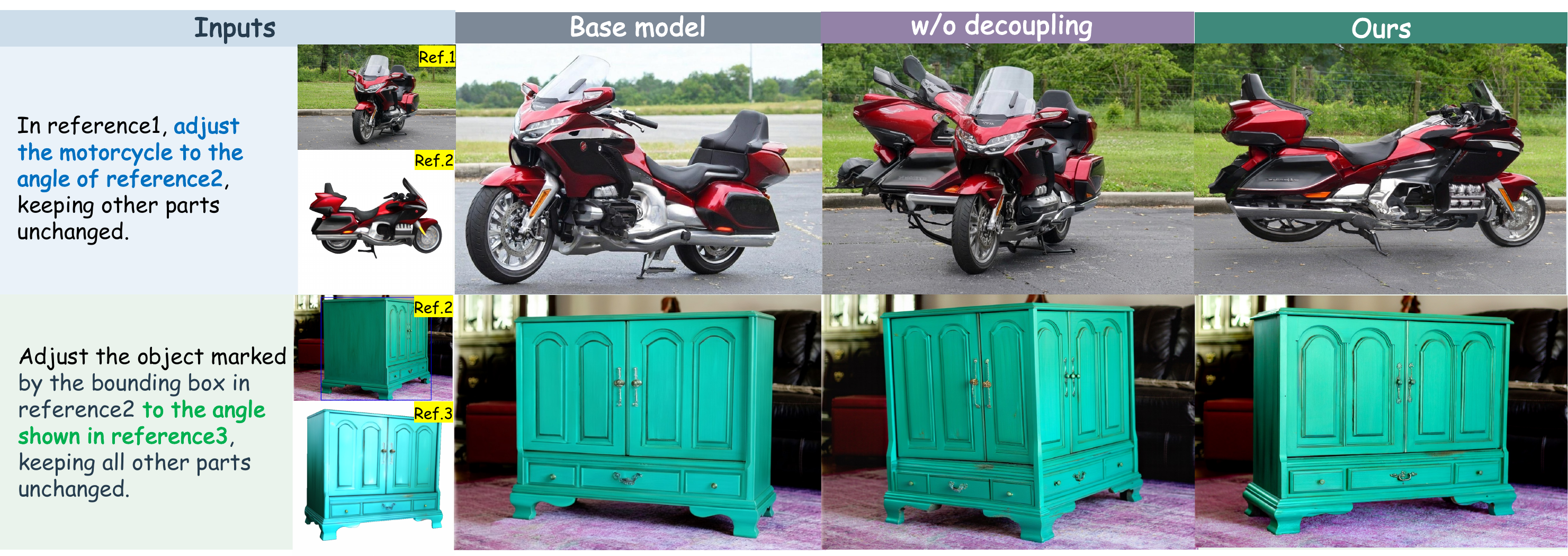}

      \caption{\textbf{Qualitative ablation study.} Our method demonstrates excellent 3D control performance.}
      \label{fig:w_o_decoupling}

\end{figure}

\textbf{Qualitative Analysis.} Fig.~\ref{fig:main_exp} and
Fig.~\ref{fig:w_o_decoupling} illustrate the failure modes that
\method removes. Existing models and baselines conditioned on a single coupled reference exhibit one of two
behaviors: they copy the flat-shaded geometry of the pose scaffold, smearing away textures,
embossed logos and printed text, or they honor the appearance reference and drift back toward the
original viewpoint, ignoring the requested rotation.
\method adopts the target orientation while transferring
high-frequency detail from the source instance, and it respects the target box in
\benchmark cases where several baselines place the object outside the requested region. Results in Tab.~\ref{tab:combined_ablation_task_metrics} further indicate the effectiveness of the appearance-geometry decoupling strategy.

\subsection{Ablation Study}
\label{sec:ablation}

\textbf{Necessity of The Fifth-stage Data Check.} As shown in Tab.~\ref{tab:combined_ablation_task_metrics},
we retrain the full model on the corpus produced \emph{before} the F5 nine-rule screening
(\textit{w/o Check}), keeping the training steps matched. Adding the F5 improves every
task category without exception: object addition $8.85{\rightarrow}9.18$, appearance editing
$8.54{\rightarrow}8.99$, pose/view editing $7.86{\rightarrow}8.57$, object replacement
$8.44{\rightarrow}8.84$, and conditional scene generation $8.17{\rightarrow}9.27$, costing $0.60$
macro points.

\begin{table}[!htbp]
\centering
\begingroup
\scriptsize
\setlength{\tabcolsep}{5.5pt}
\renewcommand{\arraystretch}{1.13}
\definecolor{UEAvgCol}{HTML}{FFDE7A}
\definecolor{UEIACol}{HTML}{EAF2FC}
\definecolor{UESCACol}{HTML}{E3F1F7}
\definecolor{UEPVACol}{HTML}{EEE9FA}
\definecolor{UEIAFCol}{HTML}{E6F3EE}
\definecolor{UENCPCol}{HTML}{EAF5E4}
\definecolor{UEIEQCol}{HTML}{F8EAF1}
\definecolor{UEDCPCol}{HTML}{F5EAF8}
\definecolor{UEDropColor}{HTML}{C83E3A}
\newcommand{\UEDropScore}[2]{  \mbox{#1\hspace{0.3pt}\raisebox{-0.55ex}{    {\color{UEDropColor}\normalsize\ensuremath{\scriptstyle\downarrow}    \tiny #2}}}}
\begin{tabular}{@{}l@{\hspace{3.5pt}}c
>{\columncolor{UEIACol}}c
>{\columncolor{UESCACol}}c
>{\columncolor{UEPVACol}}c
>{\columncolor{UEIAFCol}}c
>{\columncolor{UENCPCol}}c
>{\columncolor{UEIEQCol}}c
>{\columncolor{UEDCPCol}}c
>{\columncolor{UEAvgCol}[\tabcolsep][0pt]}c@{}}
\toprule
\textbf{Task} & \textbf{Configuration} & \textbf{IA$\uparrow$} & \textbf{SCA$\uparrow$} & \textbf{PVA$\uparrow$} & \textbf{IAF$\uparrow$} & \textbf{NCP$\uparrow$} & \textbf{IEQ$\uparrow$} & \textbf{DCP$\uparrow$} & \textbf{Avg.$\uparrow$} \\
\midrule
\multirow{4}{*}{\shortstack[l]{\tiny Object\\addition}} & Baseline & \UEDropScore{5.73}{3.52} & \UEDropScore{6.00}{3.07} & \UEDropScore{6.35}{2.81} & \UEDropScore{6.22}{2.59} & \UEDropScore{8.91}{0.64} & \UEDropScore{6.77}{2.52} & \UEDropScore{5.08}{3.99} & \UEDropScore{6.60}{2.58} \\
 & w/o Decoupling & \UEDropScore{8.53}{0.72} & \UEDropScore{8.66}{0.41} & \UEDropScore{8.47}{0.69} & \UEDropScore{8.16}{0.65} & \UEDropScore{\underline{9.32}}{0.23} & \UEDropScore{\underline{8.87}}{0.42} & \UEDropScore{8.57}{0.50} & \UEDropScore{8.65}{0.53} \\
 & w/o Check & \UEDropScore{\underline{8.80}}{0.45} & \UEDropScore{\underline{8.98}}{0.09} & \UEDropScore{\underline{8.86}}{0.30} & \UEDropScore{\underline{8.44}}{0.37} & \UEDropScore{9.23}{0.32} & \UEDropScore{8.84}{0.45} & \UEDropScore{\underline{8.88}}{0.19} & \UEDropScore{\underline{8.85}}{0.33} \\
 & \textbf{Ours} & \textbf{9.25} & \textbf{9.07} & \textbf{9.16} & \textbf{8.81} & \textbf{9.55} & \textbf{9.29} & \textbf{9.07} & \textbf{9.18} \\
\addlinespace[2pt]
\multirow{4}{*}{\shortstack[l]{\tiny Appearance\\editing}} & Baseline & \UEDropScore{5.58}{3.12} & \UEDropScore{8.76}{0.76} & \UEDropScore{8.70}{0.75} & \UEDropScore{5.39}{3.03} & \textbf{9.61} & \textbf{8.61} & -- & \UEDropScore{7.77}{1.22} \\
 & w/o Decoupling & \UEDropScore{\underline{8.55}}{0.15} & \UEDropScore{\underline{9.45}}{0.07} & \textbf{9.48} & \UEDropScore{\underline{8.18}}{0.24} & \UEDropScore{9.30}{0.22} & \UEDropScore{8.21}{0.12} & -- & \UEDropScore{\underline{8.86}}{0.13} \\
 & w/o Check & \UEDropScore{8.18}{0.52} & \UEDropScore{9.00}{0.52} & \UEDropScore{9.15}{0.30} & \UEDropScore{8.03}{0.39} & \UEDropScore{9.21}{0.31} & \UEDropScore{7.70}{0.63} & -- & \UEDropScore{8.54}{0.45} \\
 & \textbf{Ours} & \textbf{8.70} & \textbf{9.52} & \underline{9.45} & \textbf{8.42} & \underline{9.52} & \underline{8.33} & -- & \textbf{8.99} \\
\addlinespace[2pt]
\multirow{4}{*}{\shortstack[l]{\tiny Pose/view\\editing}} & Baseline & \UEDropScore{4.50}{3.94} & \UEDropScore{7.44}{1.12} & \UEDropScore{4.50}{3.94} & \UEDropScore{6.69}{1.69} & \UEDropScore{\underline{8.56}}{0.69} & \UEDropScore{7.44}{1.00} & \UEDropScore{5.64}{3.00} & \UEDropScore{6.43}{2.14} \\
 & w/o Decoupling & \UEDropScore{3.50}{4.94} & \UEDropScore{6.69}{1.87} & \UEDropScore{3.44}{5.00} & \UEDropScore{5.88}{2.50} & \UEDropScore{8.38}{0.87} & \UEDropScore{6.31}{2.13} & \UEDropScore{4.07}{4.57} & \UEDropScore{5.50}{3.07} \\
 & w/o Check & \UEDropScore{\underline{7.38}}{1.06} & \UEDropScore{\underline{7.88}}{0.68} & \UEDropScore{\underline{7.56}}{0.88} & \UEDropScore{\underline{8.00}}{0.38} & \UEDropScore{8.25}{1.00} & \UEDropScore{\underline{7.94}}{0.50} & \UEDropScore{\underline{8.07}}{0.57} & \UEDropScore{\underline{7.86}}{0.71} \\
 & \textbf{Ours} & \textbf{8.44} & \textbf{8.56} & \textbf{8.44} & \textbf{8.38} & \textbf{9.25} & \textbf{8.44} & \textbf{8.64} & \textbf{8.57} \\
\addlinespace[2pt]
\multirow{4}{*}{\shortstack[l]{\tiny Object\\replacement}} & Baseline & \UEDropScore{5.80}{2.98} & \UEDropScore{6.38}{2.59} & \UEDropScore{6.00}{2.84} & \UEDropScore{5.51}{3.12} & \UEDropScore{9.08}{0.12} & \UEDropScore{8.31}{0.41} & \UEDropScore{5.58}{3.21} & \UEDropScore{6.75}{2.09} \\
 & w/o Decoupling & \textbf{8.84} & \UEDropScore{\underline{8.87}}{0.10} & \UEDropScore{\underline{8.78}}{0.06} & \UEDropScore{\underline{8.54}}{0.09} & \textbf{9.30} & \textbf{8.87} & \textbf{8.81} & \textbf{8.85} \\
 & w/o Check & \UEDropScore{8.40}{0.38} & \UEDropScore{8.64}{0.33} & \UEDropScore{8.56}{0.28} & \UEDropScore{8.03}{0.60} & \UEDropScore{8.92}{0.28} & \UEDropScore{8.14}{0.58} & \UEDropScore{8.51}{0.28} & \UEDropScore{8.44}{0.40} \\
 & \textbf{Ours} & \underline{8.78} & \textbf{8.97} & \textbf{8.84} & \textbf{8.63} & \underline{9.20} & \underline{8.72} & \underline{8.79} & \underline{8.84} \\
\addlinespace[2pt]
\multirow{4}{*}{\shortstack[l]{\tiny Conditional\\scene\\generation}} & Baseline & \UEDropScore{6.80}{2.70} & \UEDropScore{6.10}{3.05} & \UEDropScore{6.45}{2.75} & \UEDropScore{8.45}{0.65} & -- & \UEDropScore{8.90}{0.50} & -- & \UEDropScore{7.34}{1.93} \\
 & w/o Decoupling & \UEDropScore{\underline{8.50}}{1.00} & \UEDropScore{\underline{8.90}}{0.25} & \UEDropScore{7.75}{1.45} & \UEDropScore{\underline{8.95}}{0.15} & -- & \UEDropScore{\underline{8.95}}{0.45} & -- & \UEDropScore{\underline{8.61}}{0.66} \\
 & w/o Check & \UEDropScore{8.05}{1.45} & \UEDropScore{8.30}{0.85} & \UEDropScore{\underline{7.80}}{1.40} & \UEDropScore{7.95}{1.15} & -- & \UEDropScore{8.75}{0.65} & -- & \UEDropScore{8.17}{1.10} \\
 & \textbf{Ours} & \textbf{9.50} & \textbf{9.15} & \textbf{9.20} & \textbf{9.10} & -- & \textbf{9.40} & -- & \textbf{9.27} \\
\bottomrule
\end{tabular}

\caption{\textbf{Per-task results on \benchmark}.
Baseline denotes the initialization checkpoint before fine-tuning. The w/o Decoupling and w/o Check variants are separately trained ablations.
}
\label{tab:combined_ablation_task_metrics}

\endgroup
\end{table}

\begin{figure}[!htbp]
      \centering
      \includegraphics[width=0.96\textwidth]{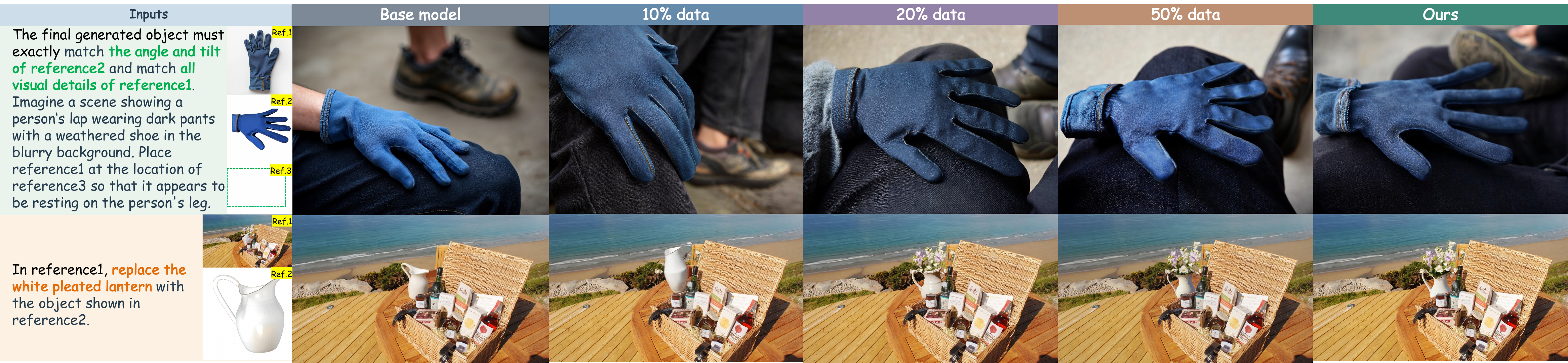}

      \caption{\textbf{Ablation results} on the impact of varying data scales on image editing performance.}
      \label{fig:data_scale}

      \end{figure}

\textbf{Data Scaling.}
We additionally train on $10\%$, $20\%$ and $50\%$ of
\dataset under the same schedule and observe a monotone improvement in editing
quality as the corpus grows (Fig.~\ref{fig:data_scale}), supporting the
scaling potential of the data engine.

\begin{figure}[!htbp]
      \centering
      \includegraphics[width=0.9\textwidth]{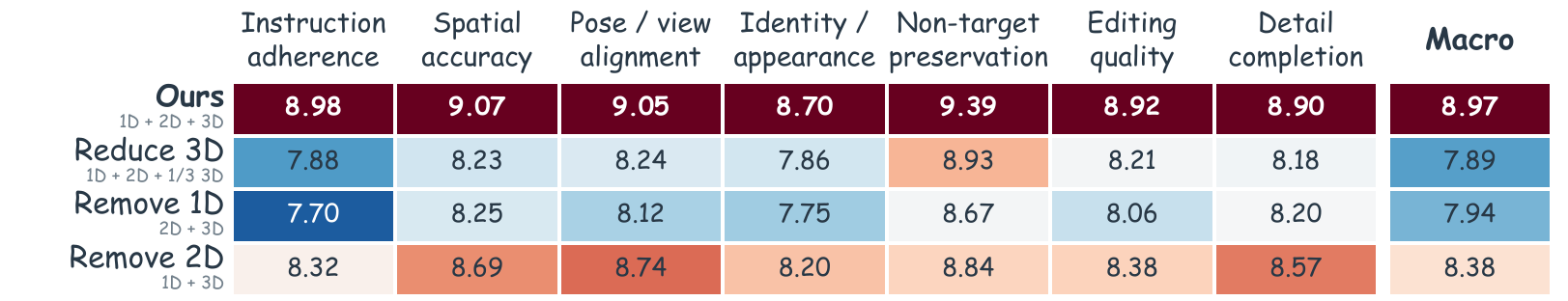}

      \caption{\textbf{Quantitative ablation study} on the effectiveness of  joint 1D/2D/3D training paradigm.}

      \label{fig:data_type}
      \end{figure}

\textbf{Joint 1D/2D/3D training.}
Fig.~\ref{fig:data_type} ablates the training mixture along the control dimensions: removing 1D
instructions ($2\text{D}{+}3\text{D}$), removing 2D spatial control
($1\text{D}{+}~3\text{D}$), and reducing the 3D
stream to one third of its volume ($1\text{D}{+}2\text{D}{+}\tfrac{1}{3}3\text{D}$). All three fall
short of the jointly trained model on the macro score ($8.97$ vs.\ $7.94$, $8.38$,  and
$7.89$). The pattern is informative rather than uniform: dropping 1D supervision hurts instruction
adherence and non-target preservation most, and dropping 2D supervision
degrades spatial compliance.
Removing any
single dimension therefore damages a different capability, which validates the unified-corpus
design over per-task specialists.

\FloatBarrier

\section{Conclusion}
In this work, we explored unified, multi-dimensional image editing across three foundational pillars: Data, Evaluation, and Methodology.
First, we established a dedicated data curation pipeline and constructed \dataset, a large 1.6 million  dataset bridging 1D textual, 2D spatial, and 3D geometric controls.
Furthermore, we established \benchmark to systematically evaluate multi-factor editing capabilities.
Moreover, we introduced \method, an MoT-based all-in-one editor featuring an appearance-geometry decoupled pipeline that leverages SAM-3D for orientation guidance while preserving high-frequency instance details from source references.
Extensive evaluations show that \method sets a new state-of-the-art, even significantly outperforming top proprietary models such as Seedream 5.0 on challenging 3D orientation tasks.

\bibliography{references}

@inproceedings{brooks2023instructpix2pix,
  title={Instructpix2pix: Learning to follow image editing instructions},
  author={Brooks, Tim and Holynski, Aleksander and Efros, Alexei A},
  booktitle={2023 IEEE/CVF Conference on Computer Vision and Pattern Recognition (CVPR)},
  pages={18392--18402},
  year={2023},
  organization={IEEE}
}

@article{zhang2023magicbrush,
  title={Magicbrush: A manually annotated dataset for instruction-guided image editing},
  author={Zhang, Kai and Mo, Lingbo and Chen, Wenhu and Sun, Huan and Su, Yu},
  journal={Advances in neural information processing systems},
  volume={36},
  pages={31428--31449},
  year={2023}
}

@inproceedings{sheynin2024emu,
  title={Emu edit: Precise image editing via recognition and generation tasks},
  author={Sheynin, Shelly and Polyak, Adam and Singer, Uriel and Kirstain, Yuval and Zohar, Amit and Ashual, Oron and Parikh, Devi and Taigman, Yaniv},
  booktitle={2024 ieee/cvf conference on computer vision and pattern recognition (cvpr)},
  pages={8871--8879},
  year={2024},
  organization={IEEE}
}

@article{zhao2024ultraedit,
  title={Ultraedit: Instruction-based fine-grained image editing at scale},
  author={Zhao, Haozhe and Ma, Xiaojian and Chen, Liang and Si, Shuzheng and Wu, Rujie and An, Kaikai and Yu, Peiyu and Zhang, Minjia and Li, Qing and Chang, Baobao},
  journal={Advances in Neural Information Processing Systems},
  volume={37},
  pages={3058--3093},
  year={2024}
}

@article{hui2024hq,
  title={Hq-edit: A high-quality dataset for instruction-based image editing},
  author={Hui, Mude and Yang, Siwei and Zhao, Bingchen and Shi, Yichun and Wang, Heng and Wang, Peng and Zhou, Yuyin and Xie, Cihang},
  journal={arXiv preprint arXiv:2404.09990},
  year={2024}
}

@inproceedings{yu2025anyedit,
  title={Anyedit: Mastering unified high-quality image editing for any idea},
  author={Yu, Qifan and Chow, Wei and Yue, Zhongqi and Pan, Kaihang and Wu, Yang and Wan, Xiaoyang and Li, Juncheng and Tang, Siliang and Zhang, Hanwang and Zhuang, Yueting},
  booktitle={2025 IEEE/CVF Conference on Computer Vision and Pattern Recognition (CVPR)},
  pages={26125--26135},
  year={2025},
  organization={IEEE}
}

@article{ye2026imgedit,
  title={Imgedit: A unified image editing dataset and benchmark},
  author={Ye, Yang and He, Xianyi and Li, Zongjian and Yuan, Shenghai and Yan, Zhiyuan and Hou, Bohan and Yuan, Li and others},
  journal={Advances in Neural Information Processing Systems},
  volume={38},
  year={2026}
}

@inproceedings{qian2026pico,
  title={Pico-banana-400k: A large-scale dataset for text-guided image editing},
  author={Qian, Yusu and Bocek-Rivele, Eli and Song, Liangchen and Tong, Jialing and Yang, Yinfei and Lu, Jiasen and Hu, Wenze and Gan, Zhe},
  booktitle={Proceedings of the IEEE/CVF Conference on Computer Vision and Pattern Recognition},
  pages={37226--37235},
  year={2026}
}

@inproceedings{luo2026editscore,
  title={Editscore: Unlocking online rl for image editing via high-fidelity reward modeling},
  author={Luo, Xin and Wang, Jiahao and Wu, Chenyuan and Xiao, Shitao and Jiang, Xiyan and Lian, Defu and Zhang, Jiajun and Liu, Dong and Liu, Zheng},
  booktitle={International Conference on Learning Representations},
  volume={2026},
  pages={33027--33056},
  year={2026}
}

@inproceedings{ju2026editverse,
  title={Editverse: Unifying image and video editing and generation with in-context learning},
  author={Ju, Xuan and Wang, Tianyu and Zhou, Yuqian and Zhang, He and Liu, Qing and Zhao, Cherry and Zhang, Zhifei and Li, Yijun and Cai, Yuanhao and Liu, Shaoteng and others},
  booktitle={International Conference on Learning Representations},
  volume={2026},
  pages={137234--137255},
  year={2026}
}

@inproceedings{yin2026reasonedit,
  title={Reasonedit: Towards reasoning-enhanced image editing models},
  author={Yin, Fukun and Liu, Shiyu and Han, Yucheng and Wang, Zhibo and Xing, Peng and Wang, Rui and Cheng, Wei and Wang, Yingming and Li, Aojie and Yin, Zixin and others},
  booktitle={Proceedings of the IEEE/CVF Conference on Computer Vision and Pattern Recognition},
  pages={23283--23293},
  year={2026}
}

@inproceedings{han2026unireditbench,
  title={Unireditbench: A unified reasoning-based image editing benchmark},
  author={Han, Feng and Wang, Yibin and Li, Chenglin and Liang, Zheming and Wang, Dianyi and Jiao, Yang and Wei, Zhipeng and Gong, Chao and Jin, Cheng and Wang, Jiaqi},
  booktitle={European Conference on Computer Vision},
  pages={287--304},
  year={2026},
  organization={Springer}
}

@article{cao2025hunyuanimage,
  title={Hunyuanimage 3.0 technical report},
  author={Cao, Siyu and Chen, Hangting and Chen, Peng and Cheng, Yiji and Cui, Yutao and Deng, Xinchi and Dong, Ying and Gong, Kipper and Gu, Tianpeng and Gu, Xiusen and others},
  journal={arXiv e-prints},
  pages={arXiv--2509},
  year={2025}
}

@article{zhao2026qwen,
  title={Qwen-image-2.0 technical report},
  author={Zhao, Bing and Wu, Chenfei and Li, Deqing and Meng, Hao and Li, Jiahao and Zhang, Jie and Zhou, Jingren and Lin, Junyang and Gao, Kaiyuan and Cao, Kuan and others},
  journal={arXiv preprint arXiv:2605.10730},
  year={2026}
}

@inproceedings{barrett2002object,
  title={Object-based image editing},
  author={Barrett, William A and Cheney, Alan S},
  booktitle={Proceedings of the 29th annual conference on Computer graphics and interactive techniques},
  pages={777--784},
  year={2002}
}

@inproceedings{feng2025dit4edit,
  title={Dit4edit: Diffusion transformer for image editing},
  author={Feng, Kunyu and Ma, Yue and Wang, Bingyuan and Qi, Chenyang and Chen, Haozhe and Chen, Qifeng and Wang, Zeyu},
  booktitle={Proceedings of the AAAI Conference on Artificial Intelligence},
  volume={39},
  pages={2969--2977},
  year={2025}
}

@article{huang2025diffusion,
  title={Diffusion model-based image editing: A survey},
  author={Huang, Yi and Huang, Jiancheng and Liu, Yifan and Yan, Mingfu and Lv, Jiaxi and Liu, Jianzhuang and Xiong, Wei and Zhang, He and Cao, Liangliang and Chen, Shifeng},
  journal={IEEE Transactions on Pattern Analysis and Machine Intelligence},
  volume={47},
  number={6},
  pages={4409--4437},
  year={2025},
  publisher={IEEE}
}

@article{bai2025qwen3,
  title={Qwen3-vl technical report},
  author={Bai, Shuai and Cai, Yuxuan and Chen, Ruizhe and Chen, Keqin and Chen, Xionghui and Cheng, Zesen and Deng, Lianghao and Ding, Wei and Gao, Chang and Ge, Chunjiang and others},
  journal={arXiv preprint arXiv:2511.21631},
  year={2025}
}

@article{diao2026sensenova,
  title={SenseNova-U1. 5: Towards Native Unified Visual Intelligence},
  author={Diao, Haiwen and Wang, Jiahao and Ding, Chenjing and Deng, Hanming and Chen, Jiangnan and Zhang, Ruixi and Wang, Ruohui and Tong, Wenwen and Fan, Xiangyu and Wang, Yubo and others},
  journal={arXiv preprint arXiv:2609.11929},
  year={2026}
}

@article{mao2026wan,
  title={Wan-image: Pushing the boundaries of generative visual intelligence},
  author={Mao, Chaojie and Xie, Chen-Wei and Zhong, Chongyang and Deng, Haoyou and Zhao, Jiaxing and Xiao, Jie and Xing, Jinbo and Zhang, Jingfeng and Zhou, Jingren and Zhang, Jingyi and others},
  journal={arXiv preprint arXiv:2604.19858},
  year={2026}
}

@article{labs2025flux,
  title={Flux. 1 kontext: Flow matching for in-context image generation and editing in latent space},
  author={Labs, Black Forest and Batifol, Stephen and Blattmann, Andreas and Boesel, Frederic and Consul, Saksham and Diagne, Cyril and Dockhorn, Tim and English, Jack and English, Zion and Esser, Patrick and others},
  journal={arXiv preprint arXiv:2506.15742},
  year={2025}
}

@article{bagel,
  title={Emerging properties in unified multimodal pretraining},
  author={Deng, Chaorui and Zhu, Deyao and Li, Kunchang and Gou, Chenhui and Li, Feng and Wang, Zeyu and Zhong, Shu and Yu, Weihao and Nie, Xiaonan and Song, Ziang and others},
  journal={arXiv preprint arXiv:2505.14683},
  year={2025}
}

@inproceedings{yenphraphai2024image,
  title={Image sculpting: Precise object editing with 3d geometry control},
  author={Yenphraphai, Jiraphon and Pan, Xichen and Liu, Sainan and Panozzo, Daniele and Xie, Saining},
  booktitle={2024 IEEE/CVF Conference on Computer Vision and Pattern Recognition (CVPR)},
  pages={4241--4251},
  year={2024},
  organization={IEEE}
}

@article{chen2025blenderfusion,
  title={Blenderfusion: 3d-grounded visual editing and generative compositing},
  author={Chen, Jiacheng and Mehran, Ramin and Jia, Xuhui and Xie, Saining and Woo, Sanghyun},
  journal={arXiv preprint arXiv:2506.17450},
  year={2025}
}

@article{qin2026scenedesigner,
  title={Scenedesigner: Controllable multi-object image generation with 9-dof pose manipulation},
  author={Qin, Zhenyuan and Shuai, Xincheng and Ding, Henghui},
  journal={Advances in Neural Information Processing Systems},
  volume={38},
  pages={133376--133400},
  year={2026}
}

@article{liang2024mixture,
  title={Mixture-of-transformers: A sparse and scalable architecture for multi-modal foundation models},
  author={Liang, Weixin and Yu, Lili and Luo, Liang and Iyer, Srinivasan and Dong, Ning and Zhou, Chunting and Ghosh, Gargi and Lewis, Mike and Yih, Wen-tau and Zettlemoyer, Luke and others},
  journal={arXiv preprint arXiv:2411.04996},
  year={2024}
}

@inproceedings{zhu2025training,
  title={Training-free geometric image editing on diffusion models},
  author={Zhu, Hanshen and Zhu, Zhen and Zhang, Kaile and Gong, Yiming and Liu, Yuliang and Bai, Xiang},
  booktitle={2025 IEEE/CVF International Conference on Computer Vision (ICCV)},
  pages={19130--19140},
  year={2025},
  organization={IEEE}
}

@inproceedings{zhang2025freeinsert,
  title={FreeInsert: Personalized Object Insertion with Geometric and Style Control},
  author={Zhang, Yuhong and Wang, Han and Wang, Yiwen and Xie, Rong and Song, Li},
  booktitle={Proceedings of the 33rd ACM International Conference on Multimedia},
  pages={10361--10369},
  year={2025}
}

@article{michel2023object,
  title={Object 3dit: Language-guided 3d-aware image editing},
  author={Michel, Oscar and Bhattad, Anand and VanderBilt, Eli and Krishna, Ranjay and Kembhavi, Aniruddha and Gupta, Tanmay},
  journal={Advances in Neural Information Processing Systems},
  volume={36},
  pages={3497--3516},
  year={2023}
}

@inproceedings{fu2024guiding,
  title={Guiding instruction-based image editing via multimodal large language models},
  author={Fu, Tsu-Jui and Hu, Wenze and Du, Xianzhi and Wang, William and Yang, Yinfei and Gan, Zhe},
  booktitle={International Conference on Learning Representations},
  volume={2024},
  pages={54820--54833},
  year={2024}
}

@inproceedings{li2023gligen,
  title={Gligen: Open-set grounded text-to-image generation},
  author={Li, Yuheng and Liu, Haotian and Wu, Qingyang and Mu, Fangzhou and Yang, Jianwei and Gao, Jianfeng and Li, Chunyuan and Lee, Yong Jae},
  booktitle={2023 IEEE/CVF Conference on Computer Vision and Pattern Recognition (CVPR)},
  pages={22511--22521},
  year={2023},
  organization={IEEE}
}

@inproceedings{zhang2023adding,
  title={Adding conditional control to text-to-image diffusion models},
  author={Zhang, Lvmin and Rao, Anyi and Agrawala, Maneesh},
  booktitle={2023 IEEE/CVF International Conference on Computer Vision (ICCV)},
  pages={3813--3824},
  year={2023},
  organization={IEEE}
}

@article{liu2025step1x,
  title={Step1x-edit: A practical framework for general image editing},
  author={Liu, Shiyu and Han, Yucheng and Xing, Peng and Yin, Fukun and Wang, Rui and Cheng, Wei and Liao, Jiaqi and Wang, Yingming and Fu, Honghao and Han, Chunrui and others},
  journal={arXiv preprint arXiv:2504.17761},
  year={2025}
}

@inproceedings{wu2026omnigen2,
  title={Omnigen2: Towards instruction-aligned multimodal generation},
  author={Wu, Chenyuan and Wang, Jiahao and Zheng, Pengfei and Yan, Ruiran and Xiao, Shitao and Luo, Xin and Wang, Yueze and Li, Wanli and Jiang, Xiyan and Liu, Yexin and others},
  booktitle={Proceedings of the IEEE/CVF Conference on Computer Vision and Pattern Recognition},
  pages={21964--21975},
  year={2026}
}

@inproceedings{chen2024geodiffusion,
  title={Geodiffusion: Text-prompted geometric control for object detection data generation},
  author={Chen, Kai and Xie, Enze and Chen, Zhe and Wang, Yibo and Hong, Lanqing and Li, Zhenguo and Yeung, Dit-Yan},
  booktitle={International Conference on Learning Representations},
  volume={2024},
  pages={8979--9001},
  year={2024}
}

@article{shi2023zero123,
  title={Zero123++: a single image to consistent multi-view diffusion base model},
  author={Shi, Ruoxi and Chen, Hansheng and Zhang, Zhuoyang and Liu, Minghua and Xu, Chao and Wei, Xinyue and Chen, Linghao and Zeng, Chong and Su, Hao},
  journal={arXiv preprint arXiv:2310.15110},
  year={2023}
}

@article{team2026firered,
  title={Firered-image-edit-1.0 technical report},
  author={Team, Super Intelligence and Qiao, Changhao and Hui, Chao and Li, Chen and Wang, Cunzheng and Song, Dejia and Zhang, Jiale and Li, Jing and Xiang, Qiang and Wang, Runqi and others},
  journal={arXiv preprint arXiv:2602.13344},
  year={2026}
}

@article{xia2025dreamomni2,
  title={Dreamomni2: Multimodal instruction-based editing and generation},
  author={Xia, Bin and Peng, Bohao and Zhang, Yuechen and Huang, Junjia and Liu, Jiyang and Li, Jingyao and Tan, Haoru and Wu, Sitong and Wang, Chengyao and Wang, Yitong and others},
  journal={arXiv preprint arXiv:2510.06679},
  year={2025}
}

@inproceedings{xiao2025omnigen,
  title={Omnigen: Unified image generation},
  author={Xiao, Shitao and Wang, Yueze and Zhou, Junjie and Yuan, Huaying and Xing, Xingrun and Yan, Ruiran and Li, Chaofan and Wang, Shuting and Huang, Tiejun and Liu, Zheng},
  booktitle={2025 IEEE/CVF Conference on Computer Vision and Pattern Recognition (CVPR)},
  pages={13294--13304},
  year={2025},
  organization={IEEE}
}

@inproceedings{chen2026sam3d,
  title={Sam 3d: 3dfy anything in images},
  author={Chen, Xingyu and Chu, Fu-Jen and Gleize, Pierre and Liang, Kevin J and Sax, Alexander and Tang, Hao and Wang, Weiyao and Guo, Michelle and Hardin, Thibaut and Li, Xiang and others},
  booktitle={Proceedings of the IEEE/CVF Conference on Computer Vision and Pattern Recognition},
  pages={7220--7232},
  year={2026}
}

@article{cai2026hidream,
  title={Hidream-o1-image: A natively unified image generative foundation model with pixel-level unified transformer},
  author={Cai, Qi and Chen, Jingwen and Gao, Chengmin and Gong, Zijian and Li, Yehao and Pan, Yingwei and Peng, Yi and Qiu, Zhaofan and Yu, Kai and Zhang, Yiheng and others},
  journal={arXiv preprint arXiv:2605.11061},
  year={2026}
}

@inproceedings{wang2025moge,
  title={Moge: Unlocking accurate monocular geometry estimation for open-domain images with optimal training supervision},
  author={Wang, Ruicheng and Xu, Sicheng and Dai, Cassie and Xiang, Jianfeng and Deng, Yu and Tong, Xin and Yang, Jiaolong},
  booktitle={2025 IEEE/CVF Conference on Computer Vision and Pattern Recognition (CVPR)},
  pages={5261--5271},
  year={2025},
  organization={IEEE}
}
\bibliographystyle{plainnat}

\end{document}